\documentclass{article}
\usepackage{graphicx}
\begin{document}

\title{Shape Estimation from Defocus Cue for Microscopy Images \\via Belief Propagation}
\author{Arnav Bhavsar\ \\ \ \\School of Computing and Electrical Engineering\\Indian Institute of Technology Mandi, India}

\date{}
\maketitle
\begin{abstract}
In recent years, the usefulness of 3D shape estimation is being realized in microscopic or close-range imaging, as the 3D information can further be used in various applications. Due to limited depth of field at such small distances, the defocus blur induced in images can provide information about the 3D shape of the object. The task of `shape from defocus' (SFD), involves the problem of estimating good quality 3D shape estimates from images with depth-dependent defocus blur. 

While the research area of SFD is quite well-established, the approaches have largely demonstrated results on objects with bulk/coarse shape variation. However, in many cases, objects studied under microscopes often involve fine/detailed structures, which have not been explicitly considered in most methods. In addition, given that, in recent years, large data volumes are typically associated with microscopy related applications, it is also important for such SFD methods to be efficient. 

In this work, we provide an indication of the usefulness of the Belief Propagation (BP) approach in addressing these concerns for SFD. BP has been known to be an efficient combinatorial optimization approach, and has been empirically demonstrated to yield good quality solutions in low-level vision problems such as image restoration, stereo disparity estimation etc. For exploiting the efficiency of BP in SFD, we assume local space-invariance of the defocus blur, which enables the application of BP in a straightforward manner. Even with such an assumption, the ability of BP to provide good quality solutions while using non-convex priors, reflects in yielding plausible shape estimates in presence of fine structures on the objects under microscopy imaging.     
\end{abstract}

\section{Introduction}
The 3D shape estimation for small or microscopic objects is an important task for various specialized applications involving microscopic structural analysis in industrial inspection or biological studies. The task, involves inferring information about the, often, fine shape variations in such objects \cite{n0,n8}. An automated approach to infer the 3D shape information directly from images, can be useful in such applications in terms of reducing manual time and effort and the perhaps, the overall system cost. Acquiring detailed images of objects at such small scale requires imaging at close range. At such close range, the depth of field for optical devices is very limited. Thus, the acquired images are affected by defocus blur. Incidentally, since the defocus blur is dependent on the depth of the points from the camera (lens), estimating the amount of blur at each pixel can provide us information about the shape of the objects. This principle is used in the task of shape from defocus (SFD), which is now a well established research area in computer vision.

Traditionally, shape from defocus (SFD) involves the processing multiple (at least 2) differently defocused images, captured from a single view, but with different camera parameters such as focal length, aperture, or distance between lens and the image plane \cite{n1,n2}. Given a set of such images, the relative defocus blur for a reference image with respect to other images can be estimated at each point of the image, thus yielding a blur-map. The spatial variation of the defocus blur across such a blur map provides an indication of the depth variation across the underlying object, which essentially provides an estimate of the object's 3D shape.

Since the pioneering work in shape from defocus \cite{n1}, various approaches have been proposed for estimating such a spatially varying blur from multiple defocused images \cite{n2,n3,n4,n5,n6,n7}, targeted towards improvement in the quality of the shape estimation. While the methods for SFD have evolved over the years in terms of achieving improved shape estimates, most of the approaches have demonstrated performance on objects with coarse shape variations \cite{n2,n4,n5}. 

On the other hand, few methods which exploit the defocus cue, and which indeed focus on achieving finely varying shape estimates \cite{n8,n9}, are typically computationally intensive, based on traditional combinatorial optimization approaches. With automation augmenting various inspection and diagnosis processes, there is often a requirement of processing large scale data for microscopy related applications. Hence, it is important for such image analysis methods (e.g. SFD) to be relatively more efficient. On the other hand, the SFD approaches in \cite{n4,n5} are arguably more efficient, than those proposed in \cite{n8,n9}. However, as mentioned earlier, these methods have demonstrated results on objects with relatively coarse shape variations. Importantly, the texture content on the scenes considered in \cite{n4,n5} is assumed to be quite high unlike \cite{n8,n9} which also demonstrates results on microscopy images with relatively low texture. Thus, considering the above discussion, there seems to be a need for relatively efficient SFD methods to explicitly demonstrates a plausible depth estimation for microscopy imaging of objects with fine shape variations.

From a methodological perspective, shape from defocus methods may be divided into two categories. While the blurring is ideally space-variant in nature (due to the 3D shape of the object), one class of approaches impose an assumption that the underlying object depth is locally constant, i.e. they assume the that in a small region, the object shape is flat (plane parallel to the camera plane). This is commonly termed as the local space-invariant (LSI) or an equi-focal approximation. As discussed later, this assumption can help in simplifying the computation, by allowing the use of local convolutions to match the corresponding regions in reference image with those in others. However, the approaches based on such an approximation are those which demonstrate results for coarse depth variations \cite{n2,n6}. On the other hand few methods have relaxed this constraint, 
\cite{n3,n8,n9}, but are typically quite computationally intensive.

In this work, we indicate that given a good combinatorial optimization approach, one can achieve shape estimation, including that for, finely varying structures, efficiently. To this end, we employ the approach of belief propagation (BP) which is one of the well established combinatorial optimization methods across various vision applications \cite{n10,n11}. In general, the belief propagation method is known to be much more efficient relative to more traditional approaches mentioned above.  We observe that the method can be applied in a straightforward manner for shape from defocus, if the cost computation is based on the LSI approximation. In spite of such an approximation, we demonstrate that it is able to yield plausible shape estimate from their microscopy images. Moreover, the examples that we consider, arguably have less texture than that considered in some well-known works on shape from defocus. Thus, based on our observations, we believe that the approach serves as a good balance between more efficient traditional local methods (but which typically yield relatively low-quality shape estimates), and the methods that can provide better shape estimates (but which are computationally more intensive).

In recent years, belief propagation is employed for shape estimation problems involving defocus blur \cite{bpdfd1,bpdfd2,bpdfdmotion1}. However, the work reported in \cite{bpdfd1} uses the BP approach as a post-processing step for smoothing the depth maps, whereas the approach in \cite{bpdfd2}, obtains an initial estimate of the depth map through BP and the emphasis is more on the deblurring task (using a different approach) than depth estimation using BP. The methods proposed in \cite{bpdfdmotion1} employs BP in a way that is proposed in this work. However, an important difference is that the work in \cite{bpdfdmotion1} also involves the motion (stereo) cue for depth estimation in addition to the defocus cue. Moreover, in all the above works, the shape variations considered are relatively coarse, similar to the ones in some other methods mentioned above. This work demonstrates that, even with only the defocus cue, an efficient approach like BP yields plausible results for depth variation.

The rest of the paper is organized as follows: In the next section we provide some preliminaries on shape from defocus (SFD). In section III, we provide a brief overview on belief propagation approach, and importantly discuss its applicability to the SFD problem. In section IV, provide and discuss our results, and conclude in section V.

\section{Shape from defocus preliminaries}
In this section, we briefly discuss the relationship between depth and blur, and the corresponding imaging model that is employed in shape from defocus. This also serves as a background to discuss the LSI assumption that we make subsequently in our BP based approach. 

For a thin lens, the radius $R$ of the defocus blur that acts on a point at depth $D$ from the camera can be related as \cite{n3} 
\begin{eqnarray}
R &=& rV\left(\frac{1}{U} -\frac{1}{D}\right)\\
  &=& rV\left(\frac{1}{F} -\frac{1}{V} -\frac{1}{D}\right)
\end{eqnarray}
where $r$ is the aperture radius, $V$ is the distance between the lens and the image plane, $F$ is the focal length. Given the above, the point-spread function (PSF) that defocuses the image pixel (corresponding to the point at depth $D$) in an hypothetical all-focused image, is popularly modeled as a Gaussian function, with the standard deviation related to $R$, and hence the depth $D$, as
\begin{eqnarray}
\sigma = \rho R = \rho rV\left(\frac{1}{F} -\frac{1}{V} -\frac{1}{D}\right)
\end{eqnarray}
where $\rho$ is a camera-dependent constant that maps the real-world distance $R$ to the $\sigma$ in terms of pixels. Thus, by estimating the blur parameter $\sigma$ at each point in the image, one can compute the corresponding depth at each point in the scene, given the camera parameters. The the estimate of $\sigma$ provides us the 3D  shape information. 

The defocus blurring operation can be expressed as a linear operation. However, as we consider general 3D scenes, the Gaussian blur at each pixel can be different depending upon its depth. Hence, the blurring operation is not space-invariant. Thus, the blurring of an image can be expressed as \cite{n3},   
\begin{equation}
g(n) = \sum_m h(n;m)f(m)
\end{equation}
where $n$ and $m$ denote pixel locations, which, for simplicity, are expressed in one dimension, $g$ denotes the blurred image, $f$ denotes the unknown focused image, and $h$ denotes the unknown blurring filter. While such a linear but space-variant modeling of defocus blurring is quite realistic for a 3D scene, we now discuss an approximation simplifies the problem as elaborated below.

\subsection{Local space-invariance in shape from defocus}
The space-variance in the above model is a result of the scene being 3D, where the depth potentially varies at each point. However, in most real scenes, the depth variation in a small (local) region is quite less or even negligible, and it is fair to assume the depth to be locally constant. Here, the local region under consideration is the support of the Gaussian blur kernel is as small as 7 - 20 pixels (in both the directions). 

Given such an approximation, for that local region, the blurring operation can be considered as space-invariant and can be expressed as a convolution \cite{n3} 
\begin{equation}
g(n) = h(n)\ast f(n)
\end{equation}
Here, the operation $\ast$ denotes convolution.

The above equation can be represented in the frequency domain as
\begin{equation}
G(\omega) = H(\omega)F(\omega)
\end{equation}
where $G$, $H$ and $F$ are Fourier transform representations of $g$, $h$, and $f$, respectively.

Based on the above approximation, one can relate two differently blurred images $g_1$ and $g_2$ images as follows,
\begin{equation}
G(\omega) = H(\omega)F(\omega)
\end{equation}
\begin{eqnarray}
G_i(\omega) &=& H_i(\omega)F(\omega)\hspace{1cm} i = 1,2\\
\Rightarrow\frac{G_2(\omega)}{G_1(\omega)} &=& \frac{H_2(\omega)}{H_1(\omega)} = H_r(\omega)\nonumber\\
\Rightarrow G_2(\omega) &=& H_r(\omega)G_1(\omega)\nonumber
\end{eqnarray}
\begin{equation}
g_2(n) = h_r(n)\ast g_1(n)
\end{equation}
Here, $h_r$ is the relative blur between $g_1$ and $g_2$, with the Gaussian blur parameter as $\sigma_r = \sqrt{\sigma_2^2 - \sigma_1^2}$. Here, it is assumed that  $g_2$ is more defocused than $g_1$ at all points.  In this case, the objective is to compute the relative blur parameter $\sigma_r$, as the estimate of the 3D shape.

Thus, one can observe that the locally space-invariant approximation allows us to relate two differently blurred images, and factors out the unknown all-focused image. Since, estimating such an all-focused image itself is a combinatorial problem, such an approximation provides a significant advantage. Moreover, it also helps in adapting the problem so as to be solved with an efficient approach such as BP, as we discuss in the next section.
 
\section{Belief propagation and its application to shape from defocus}
Having discussed the preliminaries about the relationship between depth and defocus blurring, and imaging model that is involved in SFD methods, we now turn our attention to the methodology that we employ to address the SFD task. We begin with a brief discussion on belief propagation (and its salient aspects). We then formulate the SFD problem in the BP framework. 

\subsection{Belief propagation approach}
Belief propagation has shown much promise in recent years for efficiently solving standard computer vision problems such as denoising, stereo, segmentation etc. \cite{n10,n11}. As compared to traditional combinatorial optimization problems such as simulated annealing or iterated conditional modes, the BP can be computationally much more efficient and faster in convergence. Furthermore, although not considered in this paper, BP can be made highly parallel and is well-suited for implementation on GPUs. 

The approach employed in this work is the max-product BP approach which computes the MAP estimates over a graph \cite{n11}. For images, the graph is usually a grid-graph, with nodes being pixel locations. (For the following discussion on BP, we denote nodes by $p$, $q$ and $s$, for conciseness). The max-product rule works by passing messages $m^t_{pq}(f_q)$ at time $t$ to a node $q$ from its neighbouring node $p$ of the graph as follows:

\small 
\begin{eqnarray} \label{eq:bpmess}
m^t_{pq}(f_q) = \min_{f_p}\left(D_p(f_p) + V(f_p,f_q) + \sum_{s\in \Omega(p)|q} m^{t-1}_{sp}(f_p)\right)
\end{eqnarray}
\normalsize
where $D_p(f_p)$ is the data cost at node $p$ for accepting a label $f_p$, $V(f_p,f_q)$ is the prior cost between the neighbouring nodes $p$ and $q$, and $s\in \Omega(p)|q$denotes the set of nodes in the neighbourhood of $p$, not including $q$. The message vector $m^t_{pq}$ is an $L-$dimensional vector, where $L$ is the number of labels that each node can take. This message passing is iterated for each node until convergence. At convergence, the beliefs are computed as
\begin{eqnarray} \label{eq:bpbel}
b_{q}(f_q) = D_q(f_q) + \sum_{p\in N(q)} m_{pq}(f_q)
\end{eqnarray}
The belief $b_{q}(f_q)$ at each node $q$ is a $L-$dimensional vector. The MAP solution for the label at $q$ is that $f_q$ which maximizes $b_{q}(f_q)$.

\subsection{Applicability of belief propagation to shape from defocus}
From a computational perspective, creating a blurred image, according to the LSI model can be carried out as a) multiplying each image pixel value (at, say, location $n$)  with all the values of the 2D Gaussian, whose $\sigma$ corresponds to a depth $D$, and b) Averaging the overlapping values of this scaled Gaussian centered at $n$, with the values from other scaled Gaussians at pixels neighbouring $n$. 

In this section, we formulate the depth estimation problem in a MAP framework which we solve using belief propagation (BP). To define the data cost, we relate the image in the $i^{\mbox{th}}$ view with that in the reference view. Without loss of generality, the first image is considered as the reference image ($i = 1$). For ease of explanation, we consider for now, that the reference image is modeled as a shifted and blurred version of the $i^{\mbox{th}}$ image. (This will not hold good if the reference image is more blurred than the $i^{\mbox{th}}$ image, a situation which we will discuss shortly). Thus, the relationship between the reference image and the $i^{\mbox{th}}$ image is given as above
\begin{eqnarray} \label{eq:conv}
g_1(n) = h_{r_i}(\sigma_i,n_1,n_2) \ast g_i(n)\nonumber\\
\end{eqnarray}
where $h_{ri}$ signifies the relative Gaussian blur kernel corresponding to blur parameter (standard deviation) $\sqrt{\sigma_1^2 - \sigma_i^2}$. The symbol $\ast$ denotes convolution. 

The relative blur parameter $\sqrt{\sigma_1^2 - \sigma_i^2}$ can be related to $Z$. The data cost for a particular node in the $i^{\mbox{th}}$ view is defined as
\begin{eqnarray} \label{eq:datacost}
E_{d_i}(n) = |g_1(n) - h_{r_i}(\sigma_i,n) \ast g_i(n)|
\end{eqnarray}

At this point, we note that equation (\ref{eq:datacost}) involve a convolution. However, the ideal model of equation (4) does not involve a convolution, since it models the image generation with space-variant blur. Thus, the data cost defined in equation (\ref{eq:datacost}), assumes local space-invariance which is an approximation to the actual image generation model of equation (4).

This approximation is required to make our data cost amenable to processing by the BP algorithm. The data cost in BP, for a particular label at a node, is defined as \emph{the cost incurred by that node for accepting that particular label}. For applications such as de-noising or stereo disparity estimation, the data cost at a particular node involves only the label at that node and is independent of the labels of neighbouring nodes. However, for applications involving space-variant blur, this does not hold since the blur at the neighbouring nodes also influences the observation at a particular node. This requires the current depth estimates for the neighbouring nodes. However, in its basic form, the data cost in BP, does not entertain a notion of current label estimates, and the approach simply involves finding the best label out of a set of labels that minimizes the overall cost. Defining the data cost at a node through a convolution (\ref{eq:datacost}), (i.e. using the equi-focal approximation), involves only the relative blur kernel values at the current node which in turn depends on the depth label at the current node. Thus, the data cost is rendered independent of the labels of the neighbouring nodes, allowing it to be used within the BP framework in a straightforward manner. 

The prior cost enforces a smooth solution that constrains the neighbouring nodes to have similar labels. (More fundamentally, a smoothness constraint on depth manifests from modeling the depth as a Markov random field having a joint Gibbs distribution \cite{sli}). Although, a smooth solution is preferred, we also wish to avoid over-smoothing of prominent discontinuities in the solution. Thus, the penalty for neighbouring labels being different cannot be arbitrarily high. We define the smoothness prior as a truncated absolute function which is given by
\begin{eqnarray} \label{eq:priorcost}
E_p(n,m) = \min(|Z(n)-Z(m)|,T)
\end{eqnarray}
where $n$ and $m$ are neighbouring nodes (in this case, pixel locations) in a first-order neighbourhood and $T$ is the truncation threshold to saturate the prior cost beyond a certain difference in the depth labels. 

\begin{figure}[!h]
\centering
\begin{tabular}{c c c}
\includegraphics[trim = 10mm 10mm 10mm 10mm, clip, height=100pt]{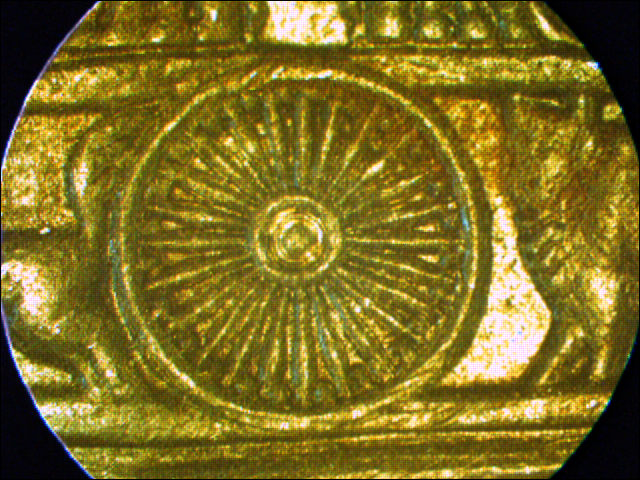}&
\includegraphics[trim = 10mm 10mm 10mm 10mm, clip, height=100pt]{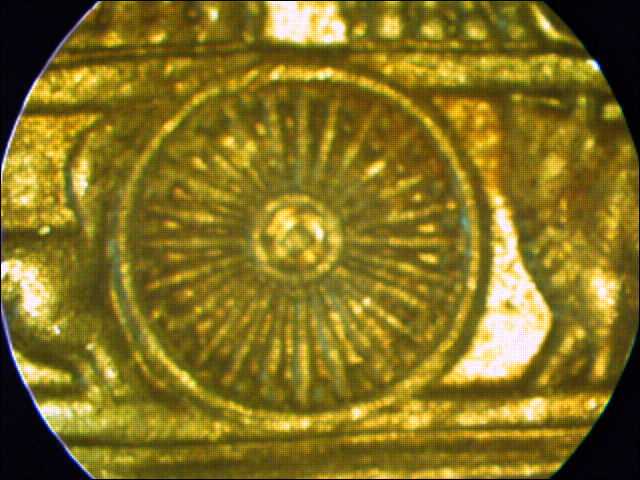}&
\includegraphics[trim = 10mm 10mm 10mm 10mm, clip, height=100pt]{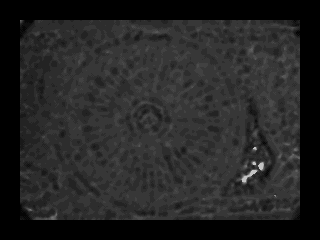}\\(a)&(b)&(c)\\
\includegraphics[trim = 10mm 10mm 10mm 10mm, clip, height=100pt]{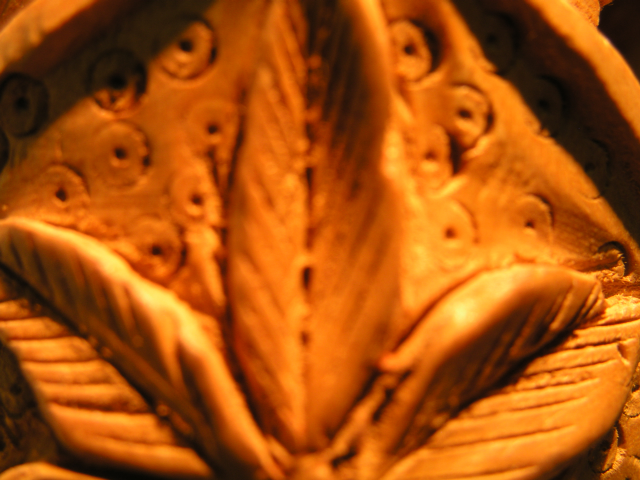}&
\includegraphics[trim = 10mm 10mm 10mm 10mm, clip, height=100pt]{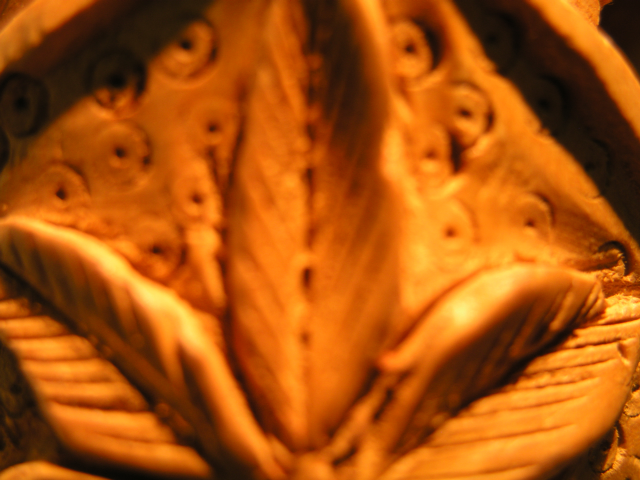}&
\includegraphics[trim = 10mm 10mm 10mm 10mm, clip, height=100pt]{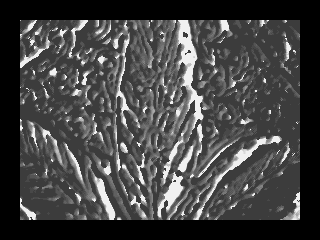}\\(d)&(e)&(f)
\end{tabular}
\caption {(a,b) Two blurred observations for the Wheel set (c) Estimated shape maps for the Wheel images. (d,e) Two blurred observations of the Flower set (f) Shape maps for the Flower images.  \label{fig:res1}}
\vspace{-0cm}
\end{figure}
\begin{figure}[!h]
\centering
\begin{tabular}{c c c c c c}
\includegraphics[trim = 10mm 10mm 10mm 10mm, clip, height=100pt]{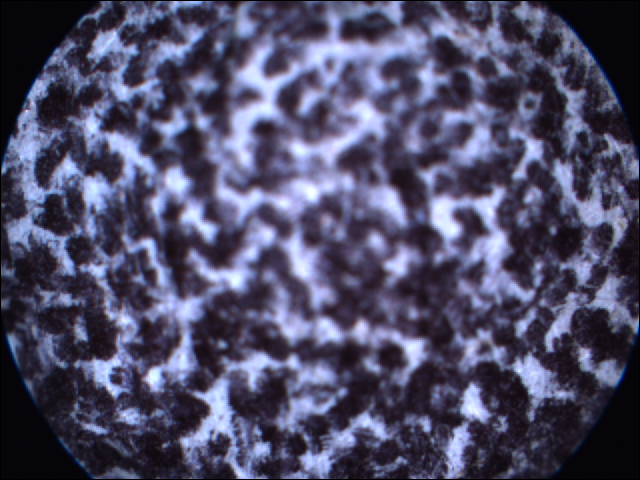} &
\includegraphics[trim = 10mm 10mm 10mm 10mm, clip, height=100pt]{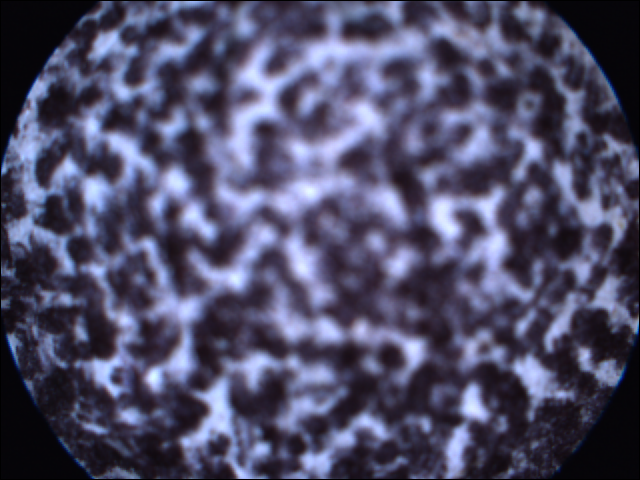} &
\includegraphics[trim = 10mm 10mm 12mm 10mm, clip, height=100pt]{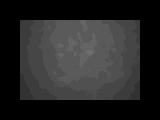}\\(a)&(b)&(c)\\
\includegraphics[trim = 10mm 10mm 10mm 10mm, clip, height=100pt]{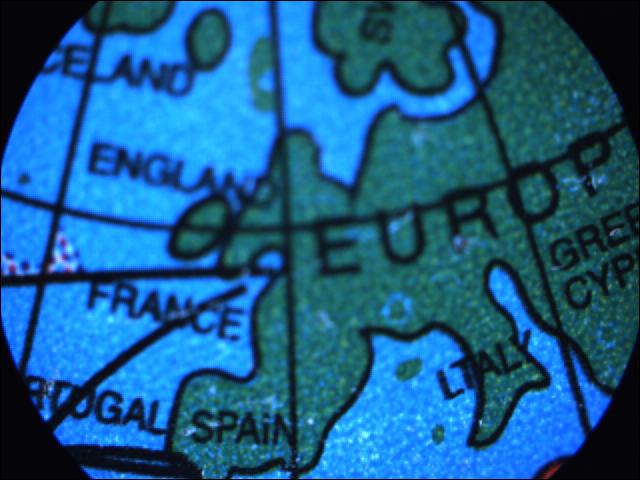} &
\includegraphics[trim = 10mm 10mm 10mm 10mm, clip, height=100pt]{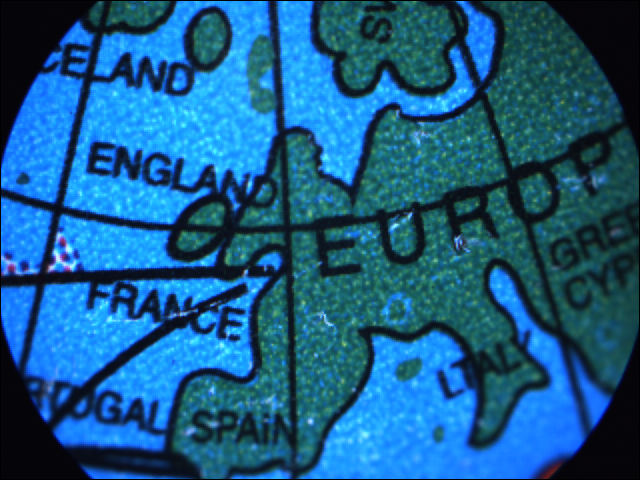} &
\includegraphics[trim = 10mm 10mm 10mm 10mm, clip, height=100pt]{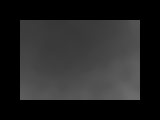}\\(d)&(e)&(f)
\end{tabular}
\caption {(a,b) Two blurred observations of the Sphere set (c) Shape maps for the Sphere images (d,e) Two blurred observations of the Globe set (f) Shape maps for the Globe images. \label{fig:res3}}
\vspace{-0cm}
\end{figure}

\section{Experimental results}
We provide some qualitative results which indicates the our BP-based method can yield plausible depth estimates across a variety of shape variations. For each experiment, we use two microscopy images of real objects, which are captured with different values of $U$ (equation (1)). In Fig. 1, we first show the results on objects which have fine variations in shape. The shape maps are shown as images, with the variations represented as gray-scale values. One can note that the shape variations indeed coincide with those perceived in the images (e.g. near spokes of the wheel (top row), or near veins of the leaves (bottom row)). However, we also note some errors in the example in the bottom row (white regions in the shape map), where there is almost no texture in the images. However, in most regions, the shape is estimated in a plausible manner even with relatively low texture.

We now demonstrate the shape estimation performance on two shapes which have smooth variation (Fig. 2). Such an experiment can also help to ascertain that the shape variation captured in the above cases, are not artifacts, but the actual fine variation of the underlying shape. We observe that the smooth shape variations are indeed captured in the shape maps of Fig. 2. The estimation in the top row shows a staircase effect in the shape estimate, as the underlying shape has a somewhat larger variation than that in the bottom case. However, in both cases, one can note that there are no major artifacts due to the image textures, and the shape variations are largely captured well.

\section{Conclusion}
In this work, we provide an indication of the usefulness of belief propagation (BP) for the task of estimating 3D shape from defocused images (SFD). We discuss and formulate the adaptation of the SFD problem to be used in the BP framework, via assuming local space invariance of defocus blur. We particularly consider the case of microscopic data, which is an important domain for the application of shape from focus. We show that the BP approach provides a efficient alternative which can also yield plausible results, especially for microscopic data, which can have finer variations in 3D shape. We believe that the reason behind its encouraging performance even under LSI approximation and low-texture is its ability to converge to good solutions efficiently.

\end{document}